\documentclass[10pt,conference,compsocconf]{IEEEtran}
\IEEEoverridecommandlockouts
\usepackage{cite}
\usepackage{amsmath,amssymb,amsfonts}
\usepackage{algorithmic}
\usepackage{graphicx}
\usepackage{textcomp}
\usepackage{xcolor}
\usepackage[capitalise]{cleveref}

\def\BibTeX{{\rm B\kern-.05em{\sc i\kern-.025em b}\kern-.08em
    T\kern-.1667em\lower.7ex\hbox{E}\kern-.125emX}}
\begin{document}

\title{Unimodal and Multimodal Sensor Fusion for Wearable Activity Recognition\\
}
\author{\IEEEauthorblockN{Hymalai Bello\IEEEauthorrefmark{1}\IEEEauthorrefmark{2}}
\IEEEauthorblockA{Email: hymalai.bello@dfki.de}
\IEEEauthorblockA{Supervised by Paul Lukowicz\IEEEauthorrefmark{1}\IEEEauthorrefmark{2}, Bo Zhou \IEEEauthorrefmark{1}\IEEEauthorrefmark{2} and Sungho Suh \IEEEauthorrefmark{1}\IEEEauthorrefmark{2}}
\IEEEauthorblockA{\IEEEauthorrefmark{1}German Research Center for Artificial Intelligence (DFKI), 67663 Kaiserslautern, Germany}
\IEEEauthorblockA{\IEEEauthorrefmark{2}Department of Computer Science, RPTU Kaiserslautern-Landau, 67663 Kaiserslautern, Germany}
}

\maketitle

\begin{abstract}
Combining different sensing modalities with multiple positions helps form a unified perception and understanding of complex situations such as human behavior.
Hence, human activity recognition (HAR) benefits from combining redundant and complementary information (Unimodal/Multimodal). 
Even so, it is not an easy task. 
It requires a multidisciplinary approach, including expertise in sensor technologies, signal processing, data fusion algorithms, and domain-specific knowledge.
This Ph.D. work employs sensing modalities such as inertial, pressure (audio and atmospheric pressure), and textile capacitive sensing for HAR. 
The scenarios explored are gesture and hand position tracking, facial and head pattern recognition, and body posture and gesture recognition. 
The selected wearable devices and sensing modalities are fully integrated with machine learning-based algorithms, some of which are implemented in the embedded device, on the edge, and tested in real-time. 

\end{abstract}

\vspace{-5pt}
\section{Problem Statement}
\vspace{-5pt}
    Human activities are highly context-dependent.
    The same activity can have a different meaning in different contexts.  
    Its complexity is proportional to the combination of parallel cues from human gestures, movements, and spoken/unspoken body language.
    Different sensor positions and multiple sensing modalities help to form a unified perception and understanding of complex situations. 
    And it takes into account the inherent variability of human behavior.
    Wearable devices are the most promising option for ubiquitous human activity recognition (HAR). 
    In contrast, vision-based systems need to be deployed in the environment, which limits their ubiquity. 
    Even so, exploiting multi-positional or multimodal information fusion is not straightforward. 
    This is mainly due to the different nature of the sensors. 
    Moreover, wearable-based HAR challenges include hardware-software-based solutions with size and user acceptance constraints. 
    Addressing this challenge requires a multidisciplinary approach, including expertise in sensor technologies, signal processing, data fusion algorithms, and domain-specific knowledge. 
    Using the most common wearable accessories on the market to deploy the HW/SW systems is one way to gain user acceptance. 
    Hence, the designs presented here are based on wristbands, goggles, headwear (helmet and sports cap), and clothing (jacket and gloves).  
    This work focuses on HW/SW co-design systems for HAR in the context of body gestures and facial and head pattern recognition (see \cref{fig:HW}).
    The state-of-the-art has mainly focused on motion capture with IMUs deployed on elastic or tight-fitting clothing \cite{butt2019inertial}.
    In the facial expression scenario, the authors of EarIo \cite{li2022eario}, present an earpiece to track facial expressions by sending 16-20 kHz sound waves around the face and tracking the deformation of the face.
    In food monitoring \cite{ChewingGlasses}, the authors proposed a chewing detection system attached to a glasses' frame and employed IMU and piezoelectric sensing modalities.

\vspace{-5pt}
\section{Research Conducted and Future Plan}

\begin{figure}
    \centering
    \includegraphics[width=\linewidth]{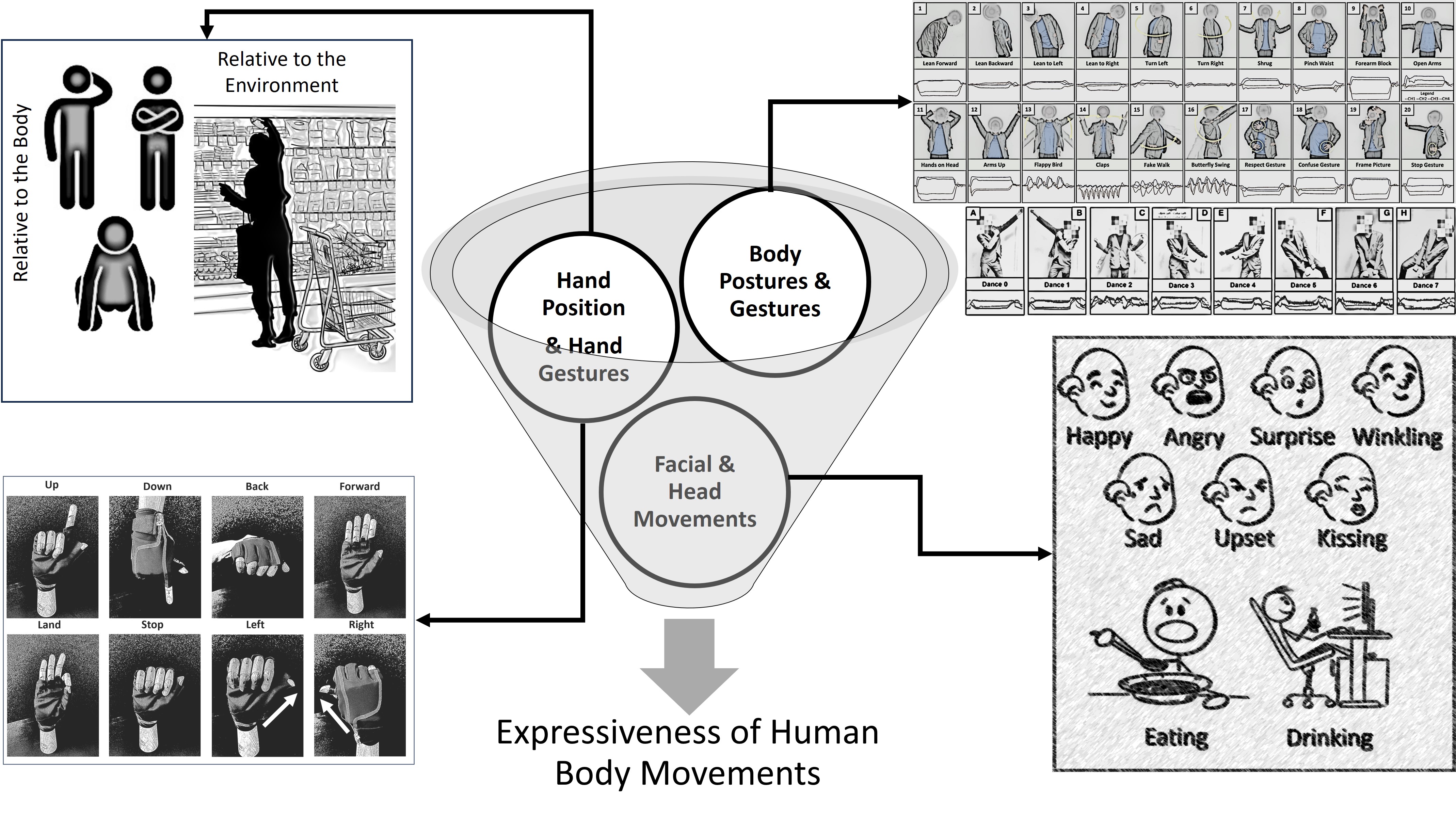}
    \caption{Simplify Diagram of the Unspoken Expressiveness of Human Body Movements with Specific Example Scenarios Studied in Thesis.}
    \label{fig:HW}
    \vspace{-10pt}
\end{figure}
This section presents an overview of the contributions this work has made to the wearable community, as shown in \cref{fig:HW}.
A comment on future direction is introduced. 

\textbf{Facial Movements:}
In \cite{bello2020facial}, a novel idea is presented to detect facial muscle movements. 
Six stethoscope microphones are deployed over the face (positions compatible with smart glasses frames) to measure sound mechanomyography.  
To our knowledge, this was the first time that passive sound was used to detect subtle facial movements such as those associated with facial expressions.
The work focuses on the fusion of unimodal and multipositional sound pressure.
As a continuation of this idea, \cite{bello2023inmyface}, presents a multimodal sensing alternative for monitoring facial muscle movement based on inertial, planar pressure, and acoustic sensors distributed in a minimally obstructive wearable accessory (sports cap). 
A modular multimodal fusion method is adopted in \cite{bello2023inmyface}. 
The fusion is based on sensor-dependent neural networks using a late fusion approach with a low memory footprint ($\leq$ 2
MB) to simplify the future deployment of the idea in wearable/embedded devices with tiny dimensions and reduce memory (4 MB to 16 MB Flash).
Facial muscle movement recognition was then combined with the detection of eating/drinking episodes in \cite{bello2023faceeat}, \cite{bello2023meciface}.
The neural network-based models were deployed on-the-edge and evaluated in real-time for both scenarios (facial and food monitoring). 
In summary, sound mechanomyography and planar-pressure mechanomyography are combined and evaluated. 
Inertial information is also merged with mechanomyography modalities with the use of wearable devices (a helmet, sports cap, and glasses). 

\textbf{Body Gestures:}
In \cite{bello2019vertical}, differential atmospheric pressure (between two barometer sensors) and radio frequency identification (RFID) synchronization are fused to recognize the vertical position of the user's hand ($\leq$30 cm) with Naive Bayes Classifier.  
Two scenarios were evaluated. 
The first scenario is shelf level recognition (6 levels) and the second is up/down position recognition around the body (Head-Chest-Feet).
A barometer is placed on the wrist with the RFID tag. 
A second barometer (as reference) is placed on a table in the first scenario (simulating an order-picking car), and in the pocket in the second scenario (simulating the position of a smartphone).
The underlying idea is that each time the user's hand approaches the reference location (order picking car or pocket), the RFID reader detects it and the stationary barometer and wrist barometer readings are synchronized.  
Wrist elevation is tracked by comparing the signal from the wrist barometer to the reference.
A comparison between unimodal, barometer only, and multimodal, barometer plus RFID synchronization, is presented.
In \cite{bello2021mocapaci}, a formal jacket (loose-fitting garment) is transformed into a wearable theremin using two off-the-shelf theremin devices and the antennas were textile capacitive antennas. 
A dictionary of 20 upper body movements was classified with deep neural networks such as 1D-LeNet, DeepConvLSTM, and Conv2D. 
The idea behind this work is that different distances between body parts can describe different postures; thus, appropriately shaped antennas embedded in garments will result in specific frequency profiles.
In \cite{bello2022move} the work is extended from unimodal, capacitive sensing only, to capacitive and RFID fusion-based synchronization.  
The start-end points of the gestures were automatically segmented by a pair of RFID tag-reader, a tag in the pocket, and a reader on the wrist.  
When the hand moves away from the pocket, the starting point is marked. 
And, when the hand is back around the pocket, the endpoint is marked. 
After RFID-based segmentation, gestures were recognized in real-time by the model running on a PC.  
Textile capacitive antennas were also deployed in sports gloves to recognize real-time and on-the-edge gestures related to drone control in The CaptAinGlove \cite{bello2023captainglove}.
The works proposed a hierarchical multimodal fusion to reduce power consumption and increase robustness against the null class, where the first stage detects movements and recognizes a non-null hand gesture using an inertial model (linear acceleration). 
Then, using a capacitive model, the second stage recognizes 8 hand gestures to control drones. 
In Body Gesture recognition, a foreseen direction is the extension of the CaptAinGlove for a realistic Smart Factory scenario. 
Pairs of CaptAinGlove are used in a real factory setting to recognize activities such as walking, opening, working, and closing factory modules to monitor the work and safety of the workers. 
Another idea is to expand the tracking of upper body movements with barometers and IMUs distributed across the body.
The idea is to use hardware built using bracelets and glasses to fuse pressure and inertial modalities in a realistic environment and with state-of-the-art fusion strategies.

\vspace{-5pt}
\section{Conclusion}

The use of unimodal, multimodal, and multi-positional sensing modalities has shown potential for robust HAR models. 
However, it is not simple.  
The goal is to have HAR models gain a deeper understanding of the expressiveness of human body movements and capture when there is a benefit from multimodal or multi-positional information.
This Ph.D. work evaluated the fusion of inertial, pressure-based (audio and atmospheric pressure), and textile capacitive sensing modalities for HAR in the context of hand position tracking, facial and head pattern recognition, and body posture and gesture recognition. 

\vspace{-5pt}
\bibliographystyle{IEEEtran}

\bibliography{References.bib}

\begin{thebibliography}{10}
\providecommand{\url}[1]{#1}
\csname url@samestyle\endcsname
\providecommand{\newblock}{\relax}
\providecommand{\bibinfo}[2]{#2}
\providecommand{\BIBentrySTDinterwordspacing}{\spaceskip=0pt\relax}
\providecommand{\BIBentryALTinterwordstretchfactor}{4}
\providecommand{\BIBentryALTinterwordspacing}{\spaceskip=\fontdimen2\font plus
\BIBentryALTinterwordstretchfactor\fontdimen3\font minus \fontdimen4\font\relax}
\providecommand{\BIBforeignlanguage}[2]{{%
\expandafter\ifx\csname l@#1\endcsname\relax
\typeout{** WARNING: IEEEtran.bst: No hyphenation pattern has been}%
\typeout{** loaded for the language `#1'. Using the pattern for}%
\typeout{** the default language instead.}%
\else
\language=\csname l@#1\endcsname
\fi
#2}}
\providecommand{\BIBdecl}{\relax}
\BIBdecl

\bibitem{butt2019inertial}
H.~T. Butt, M.~Pancholi, M.~Musahl, P.~Murthy, M.~A. Sanchez, and D.~Stricker, ``Inertial motion capture using adaptive sensor fusion and joint angle drift correction,'' in \emph{2019 22th International Conference on Information Fusion (FUSION)}.\hskip 1em plus 0.5em minus 0.4em\relax IEEE, 2019, pp. 1--8.

\bibitem{li2022eario}
K.~Li, R.~Zhang, B.~Liang, F.~Guimbreti{\`e}re, and C.~Zhang, ``Eario: A low-power acoustic sensing earable for continuously tracking detailed facial movements,'' \emph{Proceedings of the ACM on Interactive, Mobile, Wearable and Ubiquitous Technologies}, vol.~6, no.~2, pp. 1--24, 2022.

\bibitem{ChewingGlasses}
J.~Shin, S.~Lee, T.~Gong, H.~Yoon, H.~Roh, A.~Bianchi, and S.-J. Lee, ``Mydj: Sensing food intakes with an attachable on your eyeglass frame,'' in \emph{Proceedings of the 2022 CHI Conference on Human Factors in Computing Systems}, 2022, pp. 1--17.

\bibitem{bello2020facial}
H.~Bello, B.~Zhou, and P.~Lukowicz, ``Facial muscle activity recognition with reconfigurable differential stethoscope-microphones,'' \emph{Sensors}, vol.~20, no.~17, p. 4904, 2020.

\bibitem{bello2023inmyface}
H.~Bello, L.~A.~S. Marin, S.~Suh, B.~Zhou, and P.~Lukowicz, ``Inmyface: Inertial and mechanomyography-based sensor fusion for wearable facial activity recognition,'' \emph{Information Fusion}, p. 101886, 2023.

\bibitem{bello2023faceeat}
H.~Bello, S.~Suh, B.~Zhou, and P.~Lukowicz, ``Faceeat: Facial and eating activities recognition with inertial and mechanomyography fusion using a glasses-based design for real-time and on-the-edge inference,'' in \emph{Adjunct Proceedings of the 2023 ACM International Joint Conference on Pervasive and Ubiquitous Computing \& the 2023 ACM International Symposium on Wearable Computing}, 2023, pp. 199--199.

\bibitem{bello2023meciface}
H.~Bello, S.~Suh, B.~Zhou, and L.~Paul, ``Meciface: Mechanomyography and inertial fusion based glasses for edge real-time recognition of facial and eating activities,'' \emph{arXiv preprint arXiv:2306.13674}, 2023.

\bibitem{bello2019vertical}
H.~Bello, J.~Rodriguez, and P.~Lukowicz, ``Vertical hand position estimation with wearable differential barometery supported by rfid synchronization,'' in \emph{EAI International Conference on Body Area Networks}.\hskip 1em plus 0.5em minus 0.4em\relax Springer, 2019, pp. 24--33.

\bibitem{bello2021mocapaci}
H.~Bello, B.~Zhou, S.~Suh, and P.~Lukowicz, ``Mocapaci: Posture and gesture detection in loose garments using textile cables as capacitive antennas,'' in \emph{Proceedings of the 2021 ACM International Symposium on Wearable Computers}, 2021, pp. 78--83.

\bibitem{bello2022move}
H.~Bello, B.~Zhou, S.~Suh, L.~A. Sanchez~Marin, and P.~Lukowicz, ``Move with the theremin: Body posture and gesture recognition using the theremin in loose-garment with embedded textile cables as antennas,'' \emph{Frontiers in Computer Science}, vol.~4, p. 915280, 2022.

\bibitem{bello2023captainglove}
H.~Bello, S.~Suh, D.~Gei{\ss}ler, L.~S.~S. Ray, B.~Zhou, and P.~Lukowicz, ``Captainglove: Capacitive and inertial fusion-based glove for real-time on edge hand gesture recognition for drone control,'' in \emph{Adjunct Proceedings of the 2023 ACM International Joint Conference on Pervasive and Ubiquitous Computing \& the 2023 ACM International Symposium on Wearable Computing}, 2023, pp. 165--169.

\end{thebibliography}

\end{document}